\documentclass[diagnostics,article,accept,pdftex,moreauthors]{mdpi}

\firstpage{1} 
\pubvolume{1}
\issuenum{1}
\articlenumber{0}
\pubyear{2024}
\copyrightyear{2024}
\datereceived{ } 
\daterevised{ } % Comment out if no revised date
\dateaccepted{ } 
\datepublished{ } 
\hreflink{https://doi.org/} % If needed use \linebreak
\pdfoutput=1 % Uncommented for upload to arXiv.org
\usepackage{comment}
\usepackage{hhline}
 \usepackage{relsize}

\Title{Can Machine Learning  Assist in Diagnosis of Primary Immune Thrombocytopenia? A feasibility study 
}

\TitleCitation{Title}

\Author{Haroon Miah $^{1,2,\dagger}$, Dimitrios Kollias $^{3,\dagger}$, Giacinto Luca Pedone $^{2}$, Drew Provan $^{1,2}$ and Frederick Chen $^{1,2}$}

\AuthorNames{Firstname Lastname, Firstname Lastname and Firstname Lastname}

\AuthorCitation{Lastname, F.; Lastname, F.; Lastname, F.}
\address{%
$^{1}$ \quad Centre of Immunobiology, Blizard Institute, Queen Mary University of London, UK\\
$^{2}$ \quad Haematology Department, Barts Health NHS Trust, London, UK\\
$^{3}$ \quad School of Electronic Engineering \& Computer Science, Queen Mary University of London, UK}

\corres{Correspondence: d.kollias@qmul.ac.uk; frederick.chen1@nhs.net}

\firstnote{These authors contributed equally to this work.}
\abstract{
Primary Immune thrombocytopenia (ITP) is a rare autoimmune disease characterised by immune-mediated destruction of peripheral blood platelets in patients leading to low platelet counts and bleeding. The diagnosis and effective management of ITP is challenging because there is no established test to confirm the disease and no biomarker with which one can predict the response to treatment and outcome.
In this work we conduct a feasibility study to check if machine learning can be applied effectively for diagnosis of ITP using routine blood tests and demographic data in a non-acute outpatient setting.
Various ML models, including Logistic Regression, Support Vector Machine, k-Nearest Neighbor, Decision Tree and Random Forest, were applied to data from the UK Adult ITP Registry and a general hematology clinic. Two different approaches were investigated: a demographic-unaware and a demographic-aware one. 
We conduct extensive experiments to evaluate the predictive performance of  these models and approaches, as well as their bias. The results revealed that Decision Tree and Random Forest models were both superior and fair, achieving nearly perfect predictive and fairness scores, with platelet count identified as the most significant variable. Models not provided with demographic information performed better in terms of predictive accuracy but showed lower fairness score, illustrating a trade-off between predictive performance and fairness. 
}


\begin{document}

%%%%%%%%%%%%%%%%%%%%%%%%%%%%%%%%%%%%%%%%%%

\section{Introduction}
Primary Immune thrombocytopenia (ITP) is an autoimmune disease characterised by immune-mediated destruction of peripheral blood platelets in patients leading to low platelet counts and bleeding\cite{Provan2022Advances}. ITP affects approximately 6.4 per 100,000 people and although life-threatening bleeds are relatively rare, it can lead to catastrophic intracranial bleeding and death\cite{Doobaree2022ITP}. 

The diagnosis and effective management of ITP is challenging because there is no established test to confirm the disease and no biomarker with which one can predict the response to treatment and outcome. Currently, ITP diagnosis is made by exclusion of other causes of low platelet counts, and management decisions rely heavily on clinical judgment. The diagnosis therefore relies on blood tests to demonstrate a low platelet count and tests that exclude other conditions.

Artificial Intelligence (AI) encompasses the development of computer programs designed to simulate human intelligence. These programs operate on a complex framework of algorithms enabling machines to emulate human cognitive functions such as learning and problem-solving. A prominent subfield of AI, Machine Learning (ML), leverages vast datasets to identify patterns and generate predictions. What makes these algorithms unique lies in their capacity to concurrently process both linear and nonlinear variables, facilitating the recognition of intricate patterns. This capability significantly enhances their accuracy in making predictions, thus broadening their applicability across various complex scenarios.
Although ML has previously been used in healthcare to automate hospital systems, recently, it has also been utilized in the diagnosis, early detection, and monitoring of diseases \cite{tagaris2018machine,kollias2018deep,kollias2023btdnet,salpea2022medical,kollias2024domain,tagaris2017assessment,arsenos2024uncertainty,kollias2021transparent,kollias2019deep,kollias2019expression,kollias2020deep,kollias2020deeppp,kollias2021mia,kollias2022abaw,kollias2022ai,kollias2023abaww,kollias2023facernet,kollias2023multi,kollias20246th,kollias2024distribution,kolliasijcv,psaroudakis2022mixaugment,morani2024covid,karampinis2024ensuring,b300,zafeiriou2017aff}. In recent years, there have been several successful applications of AI in various medical conditions, such as the diagnosis of LA fibrillation and evaluation of prognosis in COVID-19 \cite{arsenos2022large,arsenos2023data,kollias2023ai,kollias2023deep,gerogiannis2024covid,chowdhury2023covidetector}.

%With the advent of artificial intelligence (AI), healthcare providers can leverage the power of machine learning algorithms and predictive analytics to improve the accuracy and efficiency of thrombocytopenia diagnosis and treatment.

%A machine learning (ML) system that is able to predict ITP patients using routine blood tests and demographic information before they are physically assessed in a non-acute community-facing general clinic, could streamline clinical pathways and facilitate rapid referral of ITP patients directly to a specialist haematology clinic for further confirmatory testing and therapy. This would shorten the patient journey and improve safety and efficiency.  \textcolor{red}{expand a bit?}

The potential impact of employing an ML system to predict ITP in patients using routine blood tests and demographic information includes streamlining clinical pathways, facilitating rapid referral, improving patient safety and outcomes, as well as improving efficiency.

Currently, the diagnosis of ITP often involves multiple healthcare visits and tests, which can be time-consuming and stressful for patients. The pathway typically involves initial suspicion by a GP, referral to a general clinic, and subsequent referral to a specialist, if ITP is suspected, each step requiring separate appointments and assessments. Therefore, an ML system could identify patterns or anomalies that suggest ITP enabling GPs or community clinics to make informed decisions about the necessity of specialist referrals without the need for initial physical assessments by a hematologist. This streamlined approach means that patients suspected of having ITP could bypass certain steps in the traditional pathway, reducing the time to diagnosis and treatment. Early and accurate predictions can minimize unnecessary tests and procedures, thus reducing healthcare costs and patient burden. 
Additionally, by incorporating ML predictions, community clinics can quickly identify and refer high-risk patients directly to specialist hematologists. This direct referral process avoids delays that occur when waiting for multiple consultations and non-specialist assessments.
ML models, trained on large datasets, may potentially recognize subtle patterns in blood tests that are not immediately obvious to human clinicians. This could increase the accuracy of initial assessments in community clinics, ensuring that referrals to specialists are well-founded and necessary.

Furthermore, early detection and treatment of ITP are crucial to prevent complications such as severe bleeding. ML systems can operate continuously, analyzing incoming data from routine blood tests performed for other reasons, thus potentially identifying ITP cases that might otherwise go unnoticed until symptoms worsen. ML can further help in personalizing the patient care pathway. For instance, by analyzing demographic and medical history alongside test results, ML may predict the severity of ITP or suggest the most effective treatments based on similar cases.
Finally, hospitals and clinics can better allocate their resources, including specialist time and hospital beds, by ensuring only patients with a high likelihood of having ITP are referred for specialist care. This can lead to more efficient use of healthcare resources. Reducing the number of steps in the patient journey not only improves patient experience but also reduces healthcare costs associated with multiple clinic visits and unnecessary testing.

In this work we conduct a feasibility study to check if ML can be applied effectively for diagnosis of ITP in a non-acute outpatient setting by using simple but widely available blood test results. In other words we assess if an ML model that takes as input blood test results can effectively distinguish between ITP and non-ITP patients. We utilize various widely used ML models, namely Logistic Regression (LogR), Support Vector Machine (SVM), k Nearest Neighbor (k-NN), Decision Tree (DT) and Random Forest (RF). For each model, we investigate two approaches, the demographic-unaware and the demographic-aware ones. On the former, we only provide as input to the model the blood test results, whereas on the latter (apart from the blood test results) we further provide as input to the model the patients' demographics (age, race and gender).

The blood test that we use are the routine full blood count that incorporates the platelet count, haemoglobin, red cell indices, and the differential white cell count, and biochemical screen consisting of liver and renal function tests. These are routine tests that all patients have, irrespective of suspected diagnosis when they attend a community-facing general outpatient clinic. Following diagnostic work-up, these patients are then referred on to specialist disease-specific clinics, such as our hospital’s haematology clinic for further confirmatory tests and clinical assessment. 
ITP is diagnosed based on a low platelet count (below $100 \times 10^9$/L) and exclusion of other causes of thrombocytopenia. Blood tests used routinely in outpatient clinics are able to identify low platelet counts and exclude most other causes of low platelets. 
In addition to identifying a low platelet count, the full blood count (FBC) informs of any abnormalities of the haemoglobin, white cell count, including neutrophil counts, and of red cell indices such as mean corpuscular volume (MCV).  Common causes of low platelet counts such as deficiency in vitamin B12 or folic acid, myelodysplasia, lymphoproliferative disease, bone marrow failure syndromes are associated with an abnormal FBC. An abnormal alanine aminotransferase (ALT), a commonly tested liver enzyme, and a measure of liver function, could indicate liver disease, another common cause of low platelets. %These blood tests and demographic data were therefore used as input variables for ML application. 
In this study, we used the dataset from the UK Adult ITP Registry (UKITPR). Registries are established in order to collect sufficient patient data to study the natural history and outcomes of patients, in particular of patients with rare diseases where there is insufficient experience and data available in a single institution. ITP is a rare disease, and has therefore historically not been well-studied. For this reason the UKITPR was established to collect patients’ demographic, clinical and laboratory data. The registry has recruited over 5000 patients from across the UK and from over 100 hospitals. %The eventual goal is to develop and refine a ML application that can support the diagnosis and subcategorisation of clinical patterns and risk in ITP, and predict clinical outcomes and responses to drug treatments.

We conduct extensive experiments with all ML models to discover which model  achieved the highest performance and which of the two previously mentioned approaches (i.e., demograhic-aware or demograhic-unaware) worked the best. Additionally, we conduct experiments to assess how biased each ML model is w.r.t the patients' demographics (age, race and gender). Finally, for interpretability of the achieved results, we conduct an analysis to understand the contribution of each input variable to the prediction performance of every utilized model. In this way, we can assess how important and influential each input variable is to every model. The following are the key insights from all our experiments. At first, the random forest and the decision tree ML models achieved the highest (and perfect) predictive performance across all ML models. An interesting observation is that they achieved the same performance in the two previously described approaches (demograhic-aware and demograhic-unaware ones). Secondly, these two models are the fairest among all other ML models and they are also considered fair (unlike the other ML models). Another interesting observation is that these two models achieved the same fairness score across the two approaches. Thirdly, we observe that the the low platelet count is the most important input variable that greatly (or solely) affects the decision-making of these models, a finding that is consistent with how ITP is diagnosed in the medical world (the low platelet count is also in the top-3 the most influential input variables for the rest of the ML models). Finally, across all studied ML models, the demographic-aware approach achieves  worse performance (than the demographic-unaware one), whilst it is fairer.

\section{Related Work}

Several studies have used ML to predict pre-defined outcomes in selected populations of ITP. In \cite{AN2023MLinITP}, several ML models were tested for predicting the risk of critical bleeding in a large retrospective and prospective multicentre cohort of over 3000 ITP patients, using input variables such as demographic data, comorbidities, chronicity of ITP, drugs and platelet counts. Half the cohort, constituting the retrospective data, was used for training and internal validation, and the remaining prospective data was used for testing the performance of the ML models. The best performing model in predicting critical bleeding as defined by the International Society of thrombosis and haemostasis in this study was Random Forest, which achieved  an AUC Score of 0.89; Random Forest was followed (in terms of achieved performance) by XGBoost, Light GM and logistic regression.    
In \cite{Chong2022ITPmodel},  multivariate logistic regression was utilized to predict death within 30 days of an intracranial haemorrhage (ICH) in ITP patients. Similar to the work of \cite{AN2023MLinITP}, it used a multitude of variables based on demographics, comorbidities, platelet counts at set time points from diagnosis of ITP, and drugs. Multi-centre data from 142 patients with ICH from ITP were used for training and testing ML models. The performance of their model in predicting mortality from intracranial haemorrhage was evaluated using ROC with AUC value for test cohort of 0.942.

Another study \cite{KIM2021PredictITP} had as outcome the prediction of ITP chronicity in children. In this study, a cohort of 696 single centre paediatric ITP patients was used and variables analysed at the time of ITP diagnosis were used to predict which patients will develop chronic ITP. Various ML models were tested and Random Forest was found to have the best performance in distinguishing chronic ITP from acute ITP, achieving an ROC AUC Score of 0.8. Besides demographic data and clinical features, more extensive blood tests were used in this study as variables for ML; these included presenting platelet count, immature platelet counts, platelet indices such as Mean Platelet Volume (MPV), lymphocyte count, DAT test for immune haemolysis and anti-nuclear antibody (ANA).
In a smaller study of 60 ITP patients \cite{LIU2021Blood}, ITP relapse was predicted after cessation of steroids based on the profile microbiome of the gut. Using Randon Forest, the study was able to predict relapse of ITP and response to thrombopoietin receptor agonists based on the characteristics of the microbiota, including the species of microbes. The ROC AUC values were 0.87 in distinguishing between relapse and remission of ITP.

%These studies were aimed at predicting specific defined outcomes in pre-diagnosed ITP patients, and not at the ability of ML to predict the diagnosis of ITP. Of several ML models used, RF was the best performing model. In contrast, our study interrogated various ML models to separate ITP from non-ITP in a non-acute clinic. 

%A literature review by Elshoeibi et al\cite{Yassin2023ITPReview}, demonstrated that many ML models have high predictive ability with AUC >80\% (by ROC), in predicting thrombocytopenia in the acute setting caused by sepsis or specific drugs e.g. Heparin. 
%The review suggests that AI is an effective way to predict and evaluate the outcome of thrombocytopenia patients and guide management of the patient.

%We have conducted preliminary machine learning evaluation on our dataset and have validated the ability of various ML models to recognise ITP patients from non-ITP control patients. This will serve as a basis for further development of ML models to identify characteristics to subcategorise this heterogeneous group of patients.

%%%%%%%%%%%%%%%%%%%%%%%%%%%%%%%%%%%%%%%%%%
\section{Materials and Methods}\label{section3}

\subsection{Problem Statement}

The dataset consists of $N$ data points $ (\mathbf{x}_i, y_i),$ $j= 1,\ldots,N$ with $\mathbf{x}_i \in \mathbb{R}^d$ being the input variables and their corresponding target label $y_i \in \{0,1\}$ (i.e., binary classification).
%  In our particular case, $N$ = 150, $d$ = 10 (including the 3 sensitive variables age, gender, race).
Moreover, each $\mathbf{x}_i$ is associated with the sensitive variables \footnote{A sensitive variable is a label which corresponds to a protected characteristic which we do not want to base model's decisions on.} $\mathbf{s}_i = \big < s_{\text{Age}} \text{ }, \text{ } s_{\text{Gender}} \text{ } , \text{ } s_{\text{Race}} \big>$, where $s_{\text{Age}} \in \mathbb{R}$, $s_{\text{Gender}} \in \{\text{Male}, \text{ Female}\}$ and $s_{\text{Race}} \in \{\text{White}, \text{ Black}, \text{ Asian}, \text{ Other}\}$ \footnote{We adopted a commonly accepted race classification from the U.S. Census Bureau and thus we defined the following 4 race groups (in alphabetical order): Asian, Black (or African American), White (or Caucasian) and Other (which includes Indian -or Alaska Native- and Native Hawaiian -or Other Pacific Islander-).}. In other words, these sensitive variables are the subjects' demographics. 
We develop and evaluate two approaches: a demographic-aware approach and a demographic-unaware approach. Their corresponding goals are to model $ p(y_i | \mathbf{x}_i, \mathbf{s}_i)  $ and $ p(y_i | \mathbf{x}_i)  $, respectively.

\subsection{Machine Learning Models}

Five classical Machine Learning (ML) models, namely Logistic Regression (LogR), Support Vector Machine (SVM), k-Nearest Neighbor (k-NN), Decision Tree (DT) and Random Forest (RF), were developed to diagnose ITP. 

LogR is a statistical method used for binary classification tasks in ML. It models the probability that a given input belongs to a particular class by employing the logistic function to map predicted values to probabilities between 0 and 1. This approach allows logistic regression to handle cases where the relationship between the independent variables (i.e. the input) and the dependent variable (i.e., the output) is not strictly linear. Despite its simplicity, LogR is highly effective (as it has robust performance on linearly separable data) and interpretable (as it can provide insights into feature importance and relationships), making it a popular choice for initial modeling and baseline comparisons in ML research.

SVM is a class of supervised learning algorithm that works by finding the optimal hyperplane that best separates the data into different classes. This hyperplane is determined by maximizing the margin between the nearest data points of each class, known as support vectors. SVMs are particularly effective in high-dimensional spaces and are robust to overfitting, especially in cases where the number of dimensions exceeds the number of samples. Additionally, by utilizing kernel functions, SVMs can handle non-linear classification tasks by transforming the input space into higher dimensions where a linear separation is feasible.

The k-NN algorithm is a simple yet powerful non-parametric method that operates on the principle that the classification of a data point is determined by the majority class among its k closest neighbors in the feature space, where distance metrics are used to measure proximity. It does not make any assumptions about the underlying data distribution, which makes it highly flexible and applicable to a wide range of datasets. However, k-NN's performance can be significantly affected by the choice of k and the distance metric, and it is computationally intensive for large datasets due to the need to compute distances between the query instance and all training samples. Despite these challenges, k-NN remains a widely used baseline algorithm.

DT is a versatile and widely used algorithm in ML, known for their simplicity and interpretability. It works by recursively splitting the data into subsets based on feature values, forming a tree-like structure where each internal node represents a decision based on a feature, each branch represents an outcome of that decision, and each leaf node represents a class label. One of their key advantages is their ability to model complex decision boundaries without requiring extensive data pre-processing. However, they are prone to overfitting, especially with deep trees that capture noise in the training data. Despite their limitations, DT are a fundamental tool in ML research.

RF is an ensemble learning method that enhances the performance of DTs by constructing a multitude of them during training and outputting the mode of the classes of the individual trees. This approach mitigates the overfitting typically associated with single DTs by introducing randomness through bootstrap sampling and feature selection at each split, which ensures that each tree is built from a different subset of the data and considers only a random subset of features for splitting. The aggregation of multiple trees results in a robust model that achieves high performance and generalizes well to new data. RF is highly versatile, capable of handling large datasets with higher dimensionality, and provide insights into feature importance, which is valuable for understanding underlying data structures. Their ability to handle missing values and maintain performance without extensive parameter tuning makes them a popular choice in ML research.

\subsection{Dataset}

The dataset used in this experiment originates from the United Kingdom Adult ITP Registry, hosted jointly by QMUL and Barts Health NHS Trust. %(Research Ethics Committee reference number 07/H0718/57). 
This registry, one of the largest international collections of adult primary ITP patients, encompasses detailed demographic, clinical, and genetic information from over 5000 individuals across more than 100 hospitals in the UK. With longitudinal follow-up data spanning several years, the registry serves as a comprehensive repository for understanding the characteristics and outcomes of primary ITP. 

For this experiment, we utilised data from 150 patients; 100 primary ITP patients (from the UK Adult ITP Registry) and 50 non-ITP patients,  selected from a non-acute general haematology outpatient clinic at Barts Health NHS Trust. 
In more detail, the dataset includes the demographic features: age (ranging from 29 to 106 years old), gender (Male and Female), and race (Asian, Black -or African American-, White -or Caucasian- and Other), along with key peripheral blood parameters at diagnosis: i) blood alt (liver enzyme) level; ii) blood haemaglobin level; iii)  blood neutrophil level; iv) white blood cell count; v) red blood cell count; vi) blood platelet count; vii) year of disease diagnosis. For our experiments, we employed a stratified 5-fold cross-validation strategy.

\begin{figure}[h!]
\centering
  \includegraphics[width=1.\linewidth]{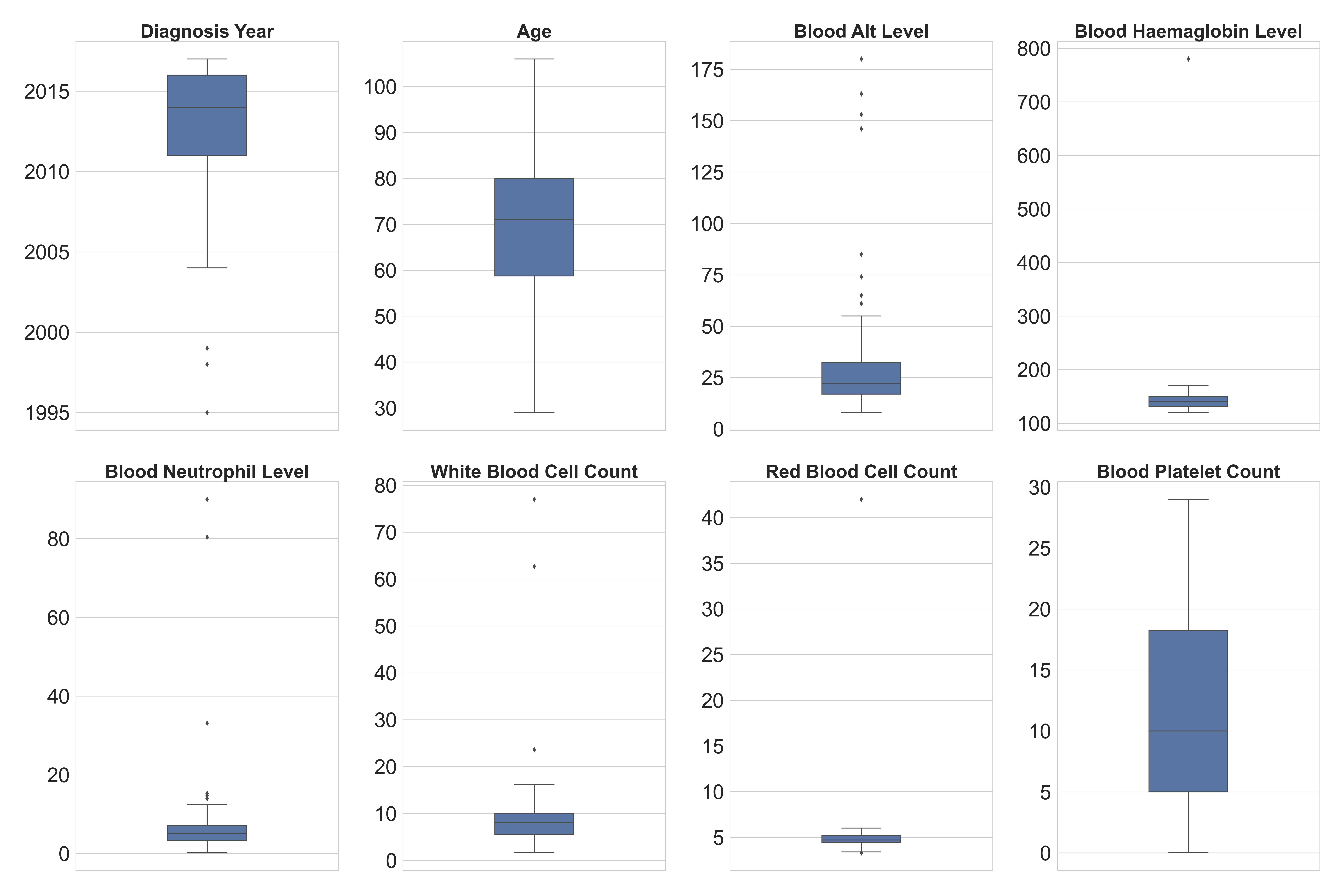}
  \caption{The boxplot of the variables across the ITP patients, namely Diagnosis Year, Age, Blood Alt Level, Blood Haemaglobin Level, Blood Neutrophil Level, White Blood Cell Count, Red Blood Cell Count and Blood Platelet Count.}
  \label{boxplot_itp}
\end{figure}

\begin{figure}[h!]
\centering
  \includegraphics[width=1.\linewidth]{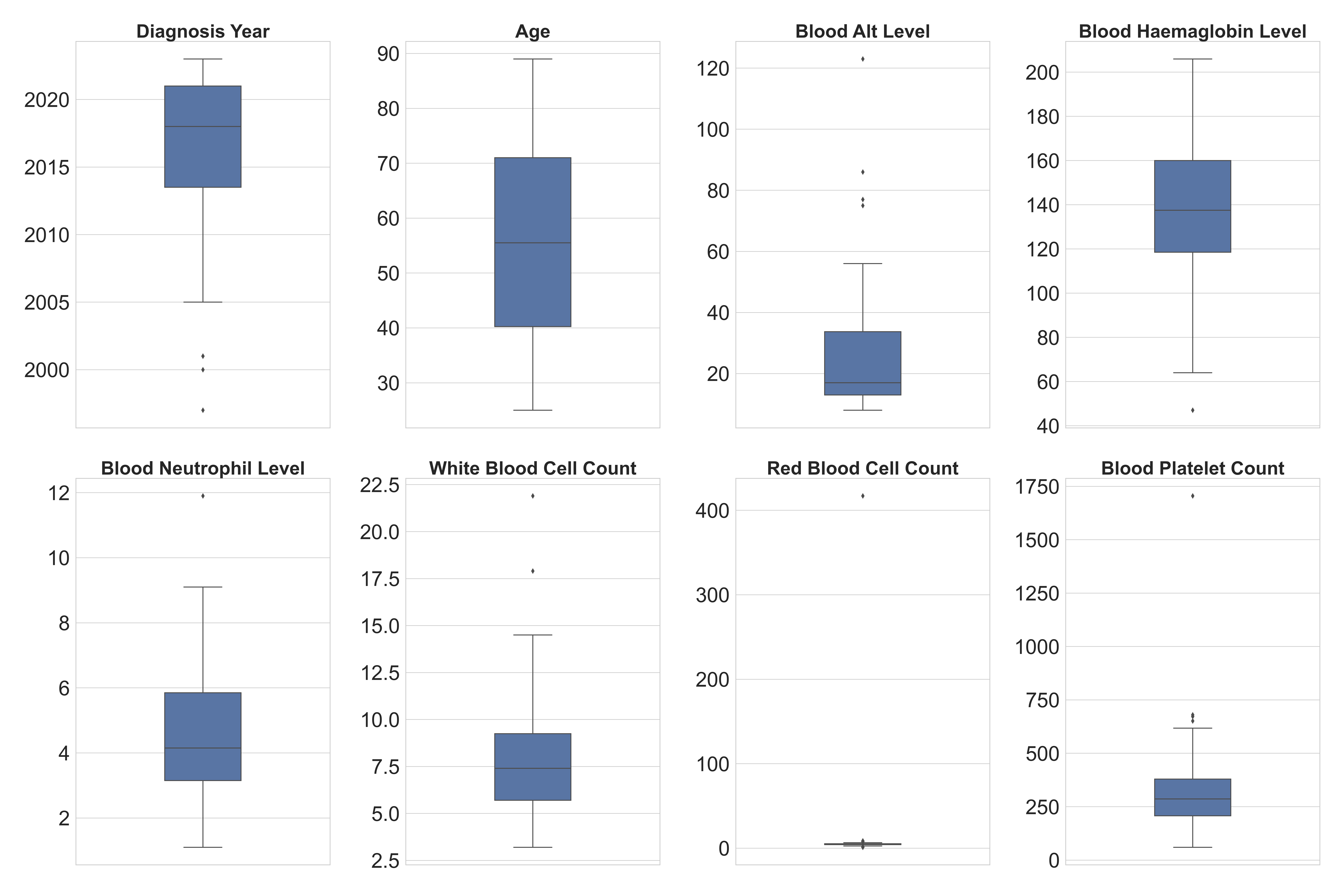}
  \caption{The boxplot of the  variables across the non-ITP patients, namely Diagnosis Year, Age, Blood Alt Level, Blood Haemaglobin Level, Blood Neutrophil Level, White Blood Cell Count, Red Blood Cell Count and Blood Platelet Count.}
  \label{boxplot_non_itp}
\end{figure}

\begin{figure}[h!]
\centering
  \includegraphics[width=0.8\linewidth]{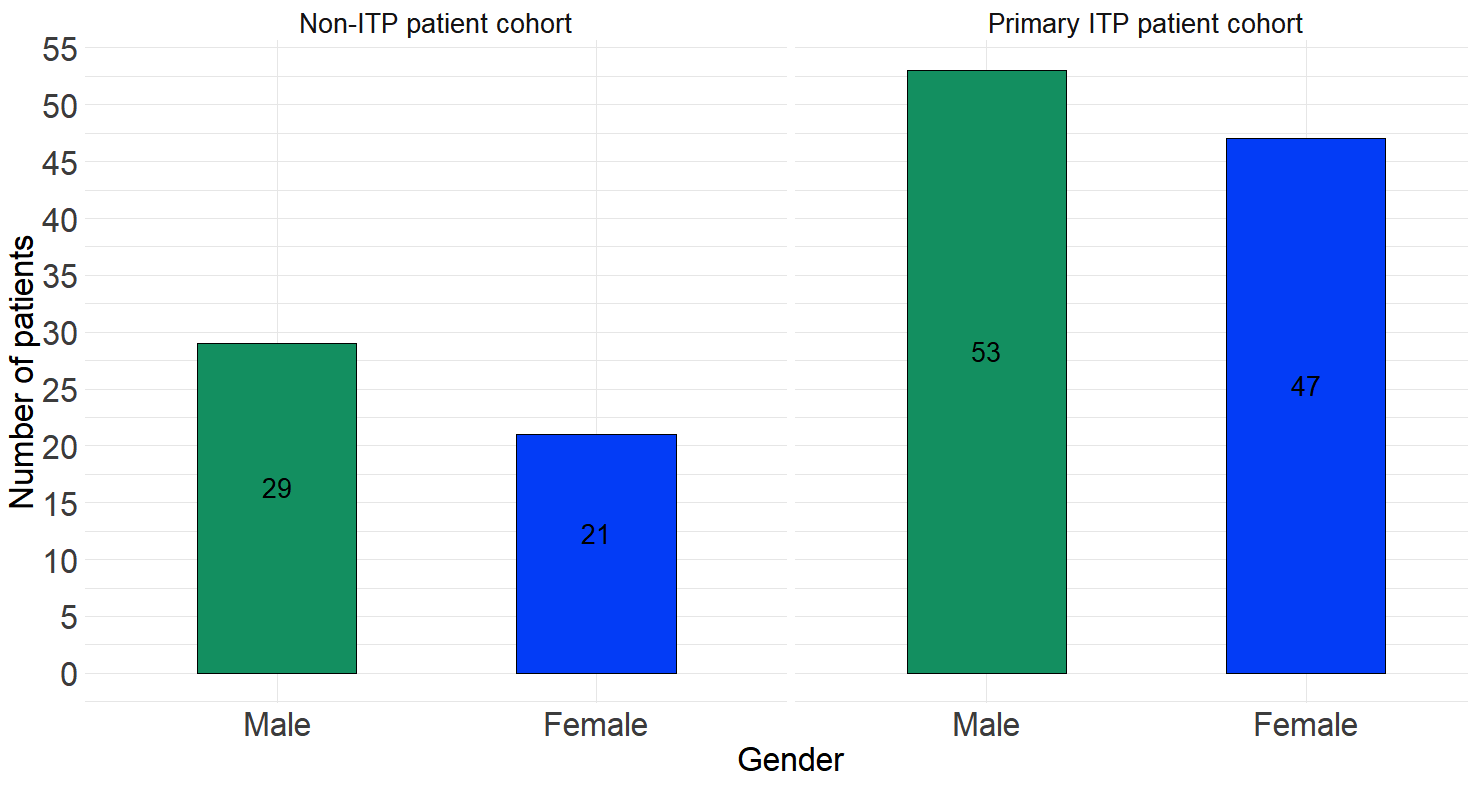}
  \caption{The gender distributions in the case of  ITP (right side) and non-ITP (left side) patients.}
  \label{boxplot_gender_itp}
\end{figure}

Figure \ref{boxplot_itp} depicts the boxplots of the numeric variables in the case of ITP patients, whereas Figure \ref{boxplot_non_itp} depicts the boxplots of these numeric variables in the case of non-ITP patients. Figure \ref{boxplot_gender_itp} presents the gender distributions (of male and female) in the case of ITP and non-ITP patients. One can see that in the ITP patient cohort the percentage of males is 53\% and in the non-ITP cohort the corresponding percentage is 58\% (i.e., 29 out of 50 patients). The distributions of male vs female patients in each cohort are quite close (53-47\% in the ITP cohort and 58-42\% in the non-ITP cohort); the distributions of males in each cohort and the ones of females in each cohort are also quite similar. Finally, Table \ref{tab:my_label} presents all numeric variables that the previously described dataset contains, along with a small description of each, their minimum, maximum, median and mean values, as well as their reference/normal ranges.

\begin{table}[h]
    \caption{Data Statistics for each numeric variable of the utilized datasets in this work. The Table also contains a small description of each variable, as well as their reference/normal ranges.}
    \label{tab:my_label} 
    \scalebox{0.8}{
    \begin{tabularx}{1.25\textwidth}{c|c|cccc|cccc|c}
    %{p{2.5cm}|p{2.5cm}|p{1cm}p{1cm}p{1cm}p{1cm}|p{1cm}
    %p{1cm}p{1cm}p{1cm}|p{1cm}}
        \toprule
         \multicolumn{2}{c|}{} & \multicolumn{4}{c|}{\textbf{Primary ITP patients}} &
         \multicolumn{4}{c|}{\textbf{Non-Primary ITP patients}} \\ [2ex]
         \textbf{Variable Name} & \textbf{Description} & \textbf{Min}  & \textbf{Max}  &  \textbf{Median} & \textbf{Mean} & 
         \textbf{Min}  &  \textbf{Max}  & \textbf{Median} &
         \textbf{Mean} & \textbf{ \begin{tabular}{@{}c@{}}Reference \\ ranges \end{tabular}} \\
         \hline
         diagnosis\_year & \begin{tabular}{@{}c@{}}  Disease diagnosis \\ year \end{tabular}  & 1995 & 2017 & - & - & 1997 & 2023 &
         - & - & N/A\\
         \hline
         age\_last\_seen & \begin{tabular}{@{}c@{}}  Age last seen \\ in clinic \end{tabular}  & 29 & 106 & 71 & 67 & 
         25 & 89 & 55.5 & 55.9 & N/A \\
         \hline
         alt & 
         \begin{tabular}{@{}c@{}}Blood ALT value \end{tabular} 
        & 8 & 180 & 22 & 31.7 & 8 & 123 & 17 & 27.2 
         & 7 - 40  \\
         \hline
         dx\_hb\_ct & 
          \begin{tabular}{@{}c@{}}Blood haemaglobin \\ count \end{tabular}
            & 120 & 780 & 141 & 148 & 47 & 206 & 137.5 & 139 
         & 115 - 180  \\
         \hline
         dx\_neutro\_ct & 
          \begin{tabular}{@{}c@{}}Blood neutrophil \\ count \end{tabular}
           & 0.17 & 90 & 5.2 & 7.4 & 1.1 & 11.9 & 4.15 & 4.15 
         & 115 - 180  \\
         \hline
         wbc\_ct &
           \begin{tabular}{@{}c@{}} White Blood Cell \\ count   \end{tabular}
           & 1.6 & 77 & 8.05 & 9.5 & 3.2 & 21.9 & 7.4 & 8 
         & 1.8 - 7.5  \\
         \hline
         rbc\_ct &  \begin{tabular}{@{}c@{}} Red Blood Cell \\ count   \end{tabular} & 3.3 & 42 & 4.71 & 5.14 & 1.09 & 417 & 4.81 & 13 & 
         3.66 - 5.54 \\
         \hline
         dx\_plt\_ct &  \begin{tabular}{@{}c@{}}Blood Platelet \\ count  \end{tabular}  & 0 & 29 & 10 & 12.3 & 60 & 1705 & 286.5 & 340 & 
         150 - 450 \\
         \bottomrule  
    \end{tabularx}
    }
\end{table}

%\textcolor{red}{describe in more detail what each input variable is + write their short names (e.g. dx neutro ct + have you included all input variables?)}%

%\textcolor{red}{i) for the 100 primary ITP patients: add some boxplots for each variable and in a table present for each variable the min, max, average and median values; ii) for the 50 controls: add some boxplots for each variable and in a table present for each variable the min, max, average and median values; iii) for the demographics do a chart for race and gender; for age show the min, max, average and median values}

%\textcolor{red}{describe the dataset: which are the input and output features; how many samples we had; mention about 5-fold cross validation and present some statistics about each fold - we could also do some EDA analysis}

\subsection{Metrics}

In the following, we present the metrics that we utilized for evaluating the performance of the ML models, as well as their fairness with respect to the sensitive variables: age, gender and race. Finally, we present the permutation feature importance technique that we utilized for measuring the importance of each individual variable in the models.

When performing a classification task, the most commonly used performance metric is the F1 Score. Generally speaking, the $F_1$ Score is a weighted average of the recall (i.e., the ability of the model-classifier to find all the positive samples) and precision (i.e., the ability of the model-classifier not to label as positive a sample that is negative). The $F_1$ Score  takes values in the range $[0,1]$; high values are desired. The $F_1$ Score is defined as:

\begin{equation} \label{f1}
F_1 = \frac{2 \times precision \times recall}{precision + recall}
\end{equation}

In our case, the performance measure is the average F1 Score (i.e., macro F1 Score) across all 2 categories (i.e., ITP patient and non-ITP patient):

\begin{equation} \label{expr}
\mathcal{P} = \frac{F_1^{\text{ITP}} + F_1^{\text{NON-ITP}}  }{2}
\end{equation}

However, the F1 Score is not sufficient in exposing differences in performance (bias) in terms of the gender, age and ethnicity sensitive variables. Therefore, we also evaluate the models using a fairness metric. %Fairness indicates whether the model is fair to the sensitive variables. 
Fairness in ML is about ensuring that the models’ decisions do not favor or discriminate against particular groups based on sensitive attributes like race, gender and age. 
There are various definitions of fairness. In this case, we use the Fairness of “Equalized Odds".

Equalized Odds is a fairness metric whose goal is to ensure that a model's performance is balanced in terms of both false positive rates (FPR) and true positive rates (TPR) across groups defined by sensitive attributes, such as race, gender, or age. TPR (also called sensitivity) is the probability that an actual positive will be correctly identified as positive. FPR is the probability that an actual negative will be wrongly identified as positive.
Equalized Odds is achieved when a model satisfies the following condition: the probability of a positive prediction given the true label should be the same across different groups. This means that both the true positive rate (TPR) and false positive rate (FPR) should be equal across these groups. 
We define the Equalized Odds as the smaller of two metrics: true positive rate ratio and false positive rate ratio. The former is the ratio between the smallest and largest of $P[\bar{y}=1 | y=1, s=\alpha]$, across all values $\alpha$ of the sensitive variable $s$, with $\bar{y}$ being the model's prediction (i.e., 0 or 1; in other words non-ITP or ITP patient), $y$ being the target label (i.e., 0 or 1). For instance, for the sensitive variable 'race' the values are 'White', 'Black', 'Asian', 'Other'. The latter is defined similarly, but for $P[\bar{y}=1 | y=0, s=\alpha]$. The Equalized Odds takes values in the range $[0,1]$; high values are desired (generally values of 90\% or more indicate fair models); the equalized odds ratio of 1 means that all groups have the same true positive, true negative, false positive, and false negative rates.

%
%Demographic parity is a fairness metric whose goal is to ensure  model’s predictions are independent of membership in a sensitive variable. In other words, demographic parity is achieved when the probability of a certain prediction is not dependent on sensitive variable membership. In our case (that we deal with binary classification), demographic parity refers to equal selection rates across the sensitive variables. Selection rate is the fraction of individuals/inputs classified as positive by the model. 
%
%We define the demographic parity as the ratio of the lowest to the highest group-level selection rates across all values of the sensitive variable. For the sensitive variable 'race' the values (i.e., the subgroups) are 'white', 'black', 'asian', 'other'.
%
%The demographic parity takes values in the range $[0,1]$; high values are desired (generally values of 90\% or more indicate fair models); a value of 1 means that all groups have the same selection rate. \textcolor{red}{grapse gia equalized odds}

%\textcolor{red}{Demographic Parity is a fairness metric that measures whether the outcomes of a model are independent of the sensitive demographic attributes. A higher Demographic Parity Score implies that the model decisions are more uniform across different demographic groups, reflecting less bias.}

Finally, we present the permutation feature importance technique that we  utilized for measuring the importance of each individual variable in the models. Permutation feature importance is a powerful tool for variable selection and model interpretation, helping in identifying variables that significantly impact the model's predictive power, and those that do not contribute meaningfully and can potentially be removed without loss of performance. It is applicable to any model and is particularly useful because it is model agnostic—meaning it does not depend on the model internals and can be used with any ML model.

Permutation feature importance involves the following four steps. After a model is trained and evaluated, we select one variable in the dataset and permute (i.e., shuffle) its values among the data points. This disruption breaks the relationship between the variable and the target, effectively making the feature irrelevant. Next, with the permuted feature, we evaluate the model  using the same performance metric that was used when it was originally trained. Because the association between the variable and the outcome has been disrupted, the model's performance is expected to degrade if the variable was important. Following that, the importance of the variable is determined by the change in the model’s performance metric caused by shuffling the variable's values. A significant decrease in performance indicates that the model relied heavily on that variable for making predictions. Conversely, a small or no change suggests that the variable was not very important for the model's predictions. Finally, we perform this process for each variable in the dataset to gauge the relative importance of all features.

\subsection{Pre-processing and Implementation Details}

% add details of hyperparameters of ML models + scikit learn + sklearn
% %  In our particular case, $N$ = 150, $d$ = 10 (including the 3 sensitive variables age, gender, race) i pes d=7 + 3 sensitive attributes

In terms of pre-processing, we applied min-max normalization to each input variable independently. In terms of implementation details, we experiments with different hyperparameters for the ML models. In the case of SVM, the kernels that we have utlized in this work are the linear (LN), the radial basis function (RBF) and the polynomial with degree two, three and four (P2, P3 and P4, respectively). In the case of k-NN, we utilized 1, 2, 4, 8 and 12 nearest neighbours (in other words we used k = 1, 2, 4, 8 and 12). In the case of permutation feature importance,  we selected 10 as the number of times to permute a variable. The scikit-learn library was utlized for our implementations.

%%%%%%%%%%%%%%%%%%%%%%%%%%%%%%%%%%%%%%%%%%
\section{Experimental Results}
In the following we provide an extensive experimental study in which we compare the performance (in terms of the F1 Score) of all utilized ML models, as well as the performance of the demographic-aware and the demographic-unaware approaches for each ML model. We also assess how biased each model is, by comparing their performance in terms of the Equalized Odds fairness metric, as well as how more or less biased the demographic-aware approach is to the demographic-unaware one. Finally, we present the permutation feature importance technique's results for each ML model and for both approaches.

\subsection{Demographic-aware vs Demographic-unaware Performance Comparison across ML Models} %: Demographic-aware vs -unaware approach}

Table \ref{perf_comp} presents a performance comparison in terms of the F1 Score (shown in \%), across all folds, between the demographic-aware and  demographic-unaware approaches of multiple machine learning methods (described in Section \ref{section3}). 

It can be observed that overall, the Random Forest (RF) method outperformed all others, achieving an 100\% F1 Score (i.e., perfect Score) across all folds in both approaches, making it the most robust method. 
The second best performing method was the Decision Tree (DT) that also performed exceptionally well with near-perfect F1 Scores. In particular, DT achieved an average F1 across all folds of 99.2\% in both approaches, achieving perfect (i.e., 100\%) F1 Scores on 4 out of 5 folds.
One can also see in Table \ref{perf_comp} that, for the k-NN method, best results have been achieved when the number of neighbors (k) was two, in both the demographic-aware and demographic-unaware approaches; overall, the higher the number of k considered, the worst the method's performance was getting in both approaches.
Finally, for the SVM Demographic-aware Approach, PL kernels outperformed the standard RBF and LN kernels; what is more, the higher the order of the PL, the higher the performance was achieved. The achieved results for the SVM Demographic-unaware Approach were more mixed (i.e., best performance was achieved when using a second order PL kernel; all PL kernels outperformed the LN kernel).

Finally, Table \ref{perf_comp} shows that the demographic-unaware approach tends to perform better or equally across all machine learning models compared to the demographic-aware approach. Only the RF and DT methods achieved the same performance across the two approaches, which in turn means that they are more robust towards the inclusion or exclusion of demographic information; all other ML models achieved a better performance when demographic information was not included. 
%\textcolor{red}{which in turn means that the demographic information made them more biased (?) and confused them + check feature importance  + kanto link me ta next sections; we show that the fairness was better without demographics + sta important features i demographic aware approach eixe ta race se ola ta models ektos apo ta best performing dt+rf }
When demographic information is included as input variables, models  overfit these variables at the expense of others that are more generalizable. This degrades the performance since the demographic variables do not have a strong, consistent predictive relationship with the outcome. What is more, demographic variables distract the model from focusing on more predictive variables, reducing its overall performance. The improvement in models without these features suggests that other variables might capture necessary information more effectively for the tasks at hand.
The higher F1 Scores observed in the demographic-unaware approach suggest that, in this specific context, excluding demographic information allows models to generalize better and focus on features that directly impact the outcome, leading to higher overall performance.

%
%\textcolor{red}{We observed that although the accuracies achieved by the three models are relatively similar, their abilities in mitigating bias are notably different. The attribute-aware approach utilizes the attribute information to allow the model to classify the expressions according to the subjects sharing similar attributes, rather than drawing information from the whole dataset}

\begin{table}[H] 
\caption{Performance comparison in terms of F1 Score (in \%) between the demographic-aware and demographic-unaware approaches across Logistic Regression (LogR), Support Vector Machines (SVM) with different kernels (radial basis function, linear and polynomial of various orders), k-Nearest Neighbors (with different 'k'), Decision Tree (DT) and Random Forest (RF). High values of F1 Score are desired.
 \label{perf_comp}}
\scalebox{0.9}{
\begin{tabularx}{1.12\textwidth}{c|cccccc||cccccc}
\toprule
\multicolumn{1}{c|}{\textbf{Method}} & \multicolumn{12}{c}{\textbf{F1 Score}}   \\
\multicolumn{1}{c|}{\textbf{}} & \multicolumn{6}{c||}{\textbf{Demographic-aware Approach}} & \multicolumn{6}{c}{\textbf{Demographic-unaware Approach}}   
\\
	& \textbf{Fld 1} & \textbf{Fld 2} & \textbf{Fld 3} & \textbf{Fld 4} & \textbf{Fld 5} & \textbf{Mean} & \textbf{Fld 1} & \textbf{Fld 2} & \textbf{Fld 3} & \textbf{Fld 4} & \textbf{Fld 5} & \textbf{Mean}\\
\hline
LogR & 72.1 & 65.9 & 77.8 & 68.9 & 79.5 & 72.8    & 75.1 &61.3 & 80.8 & 75.1 & 75.1 & 73.5\\
\hline 
SVM-RBF & 72.1 & 65.9 & 72.1 & 74.4 & 79.5 &  72.8    & 87.7 & 92.1 & 87.7 & 88.5 & 96.2 & 90.4 \\
SVM-LN & 83 & 65.9 & 87.7 & 84.1 & 84.1   & 81      & 77.1 & 77.1 & 82.8 & 88 & 88 & 82.6   \\
SVM-P2 & 83 & 71.3 & 83 & 84.1 & 88.5   & 82     & \textit{83} & \textit{96.2} & \textit{87.7} & \textit{92.1} & \textit{96.2} & \textit{91}\\
SVM-P3 & 83 & 85 & 83 & 96.3 & 96.3 & 88.7       & 83.5 & 88.2 & 88.1 & 88.2 & 96.7 & 88.9\\
\textit{SVM-P4} & \textit{83} & \textit{89} & \textit{83} & \textit{92.8} & \textit{96.3} & \textit{88.8}     & 83 & 87.7 & 92.1 & 87.7 & 96.2 & 89.3 \\
\hline 
1-NN & 83 & 85 & 83 & 82.4 & 89 & 84.5   & 96.2 & 100 & 92.8 & 89 & 96.2 & 94.8  \\
\textit{2-NN} & \textit{83} & \textit{89} & \textit{92.1} & \textit{79.2} & \textit{85.7} & \textit{85.8}       &  \textit{96.2} &  \textit{96.2} &  \textit{96.2} &  \textit{92.5} &  \textit{96.2} &  \textit{95.4}    \\
4-NN & 83 & 80.8 & 87.7 & 84.1 & 88.5 & 84.8    & 87.7 & 92.1 & 96.2 & 84.1 & 96.2 & 91.2\\
8-NN &77.8 & 83 & 83 & 71.3 & 76.2 & 78.2     & 92.1 & 92.1 & 96.2 & 79.5 & 96.2 & 91\\
12-NN & 65.6 & 58.3 & 72.1 & 62.7 & 74.3 & 66.6     & 83 & 92.1 & 87.7 & 87.7 & 92.1 & 88.5\\
\hline 
\underline{DT} & \underline{100} & \underline{96.2} & \underline{100} & \underline{100} & \underline{100} & \underline{99.2}   &  \underline{100} & \underline{96.2} & \underline{100} & \underline{100} & \underline{100} & \underline{99.2}\\
\hline 
\textbf{RF} & \textbf{100} & \textbf{100} & \textbf{100} & \textbf{100} & \textbf{100} & \textbf{100}   & \textbf{100} & \textbf{100} & \textbf{100} & \textbf{100} & \textbf{100} & \textbf{100}\\
\bottomrule
\end{tabularx}
}
\end{table}

\subsection{Demographic-aware vs Demographic-unaware Fairness Comparison across ML Models}

%mean Equalized Odds score across all folds

Table \ref{fair_comp} presents a fairness comparison -for the  sensitive variables gender, race and ethnicity- in terms of the Equalized Odds Score (presented in \%), across all folds, between the demographic-aware and  demographic-unaware approaches of multiple machine learning methods (described in Section \ref{section3}). The following observations stand for all sensitive variables (gender, race and ethnicity).

It can be observed that overall, RF and DT models outperformed all others in fairness metric; in other words, these models were the fairest among all ML methods. Furthermore, they were fair models as they achieved 100\% and 90\% Equalized Odds Scores, respectively (any model with $\ge 90\%$ Equalized Odds Scores is considered fair). 
From Table \ref{fair_comp}, one can also note that RF and DT have identical fairness Scores in both demographic-aware and demographic-unaware approaches, suggesting that these models' fairness does not  change with demographic information. It is worth noting that RF and DT achieved the best and second best performance in terms of F1 Score, as can be seen in Table \ref{perf_comp}. Therefore these models achieved both the best F1 Score performance and the best  Equalized Odds fairness performance. From Table \ref{fair_comp}, one can also notice that the rest of ML models are not fair as they achieved Equalized Odds Scores of  $\le 70\%$. These models did not also achieve the highest F1 Score performance according to Table \ref{perf_comp}. 

Table \ref{fair_comp} further shows that, among the SVM models, SVM-LN is the fairest model and this observation is the same in both the demographic-aware and demographic-unaware approaches. For the polynomial SVM models, for the demographic-aware approach,  one can see that the higher the degree of the polynomial, the fairer the performance is; whereas, for the demographic-unaware approach, the fairness score is the same no matter what the degree of the polynomial is.
Additionally, Table \ref{fair_comp} shows that, among the k-NN models, 2-NN is the fairest in the demographic-unaware approach and this model was also the best performing in terms of F1 Score (shown in Table \ref{perf_comp}). Generally, we see that, for the demographic-aware approach, the higher the k in k-NN (i.e., the more neighbors its has), the fairer the model is.

Finally, Table \ref{fair_comp} shows that the demographic-aware approach tends to be fairer or equally fair (w.r.t. Equalized Odds Score), across all ML models, compared to the demographic-unaware approach. Only the RF and DT methods achieved the same fairness score across the two approaches (which means that they are more robust towards the inclusion or exclusion of demographic information); all other ML models achieved a higher fairness score  when demographic information was included.
This observation is the opposite than the one we made when comparing their performance w.r.t. the F1 Score (from Table \ref{perf_comp}). In other words, the demographic-aware approach is fairer but exhibits a worse performance compared to the demographic-unaware approach (RF and DT are the exception as these models achieved both the highest performance and the highest fairness scores, whilst having the same performance and fairness score between the demographic-aware and  demographic-unaware approaches). This may seem contradictory or bizarre but in fact it is not.

In the demographic-aware approach, the sensitive variables were fed as input to the model which in turn paid particular attention to them and thought they were quite important and thus became fairer towards them. This is illustrated and proved in the next subsection that each ML model's input feature importance is shown. However, in parallel, in the demographic-aware approach, the models achieved a worst performance. 

It is known in the ML community the existence of a tradeoff between model performance and fairness \cite{kleinberg2016inherent,ma2022tradeoff,menon2018cost} which is primarily due to inherent differences in data distributions, historical biases present in the training data, and the conflicting goals of optimizing for performance-accuracy versus equity. 
Groups within a dataset might exhibit different statistical patterns due to a variety of factors. When a model aims for high performance, it leverages these patterns to make predictions. However, fairness considerations often require the model to treat different groups similarly, even if their underlying distributions differ. Adjusting the model to ignore these differences (to achieve fairness) can lead to a reduction in performance because the model is no longer fully optimized according to the natural distributions of the data. Such adjustments can force the model to choose less optimal solutions from a purely statistical perspective in order to enhance equity. This approach might mean rejecting the statistically "best" model in favor of one that is less accurate overall but more balanced in terms of demographic impact.

\begin{table}[H] 
\caption{Sensitive variables (Gender, Race, Age): Fairness comparison in terms of Equalized Odds (in \%) between the demographic-aware and demographic-unaware approaches across Logistic Regression (LogR), Support Vector Machines (SVM) with different kernels (radial basis function, linear and polynomial of various orders), k-Nearest Neighbors (with different 'k'), Decision Tree (DT) and Random Forest (RF). High values of Equalized Odds are desired (generally values of 90\% or more indicate fair models).
 \label{fair_comp}}
\scalebox{1.}{
\begin{tabularx}{.88\textwidth}{c|ccc||ccc}
\toprule
\multicolumn{1}{c|}{\textbf{Method}} & \multicolumn{6}{c}{\textbf{Equalized Odds Score}}   \\
\multicolumn{1}{c|}{\textbf{}} & \multicolumn{3}{c||}{\textbf{Demographic-aware Approach}} & \multicolumn{3}{c}{\textbf{Demographic-unaware Approach}}   
\\
	& \textbf{Gender} & \textbf{Race} & \textbf{Age} &  \textbf{Gender} & \textbf{Race} & \textbf{Age}\\
%\midrule
\hline
LogR &   61.1 &  52.2  & 47.9  & 49.8  & 43.1   &  37.5 \\
\hline 
SVM-RBF &  52.1   &  48.1  & 45  & 30  &  37.9  & 39 \\
SVM-LN & \textit{62.2} & \textit{54.6}   &  \textit{62.3}  & \textit{43.6}  &  \textit{54.1}  & \textit{57.4} \\
SVM-P2  & 34.6 &  36.1  & 40.4   & 22.4  &  26.3  & 28 \\
SVM-P3 & 45.8 &  52.1  &  55.1  & 22.4  & 26.3   & 28 \\
SVM-P4 & 46.2 &  52.8  &  45.7  & 22.4  & 26.3   &  28 \\
\hline 
1-NN & 20.1 &  21.2  &  24  & 12.2 & 13.6   & 15.2   \\
2-NN &  36 & 33.4   &  29.5  & \textit{31.6} & \textit{32.4} & \textit{29.1}   \\
4-NN &  42 &  38.9  &  33.3  & 28.4  &  31.5  & 28.2 \\
8-NN & 56.7 &  48.6  & 37.5   & 19.7  & 26.7   & 27.1 \\
12-NN & \textit{69.6} & \textit{55.4}   &  \textit{42.4}  & 16.4  &   18.9 & 23.8 \\
\hline 
\underline{DT} &\underline{ 90} & \underline{ 90}  &  \underline{ 90}   & \underline{90}  & \underline{ 90} &\underline{90}   \\
\hline 
\textbf{RF} & \textbf{100} & \textbf{100}   &  \textbf{100}  & \textbf{100}  &  \textbf{100}  & \textbf{100} \\
\bottomrule
\end{tabularx}
}
\end{table}

\subsubsection{Feature Importance of Demographic-aware vs Demographic-unaware ML Models}

\begin{figure}[h!]
\centering
  \includegraphics[width=.49\linewidth]{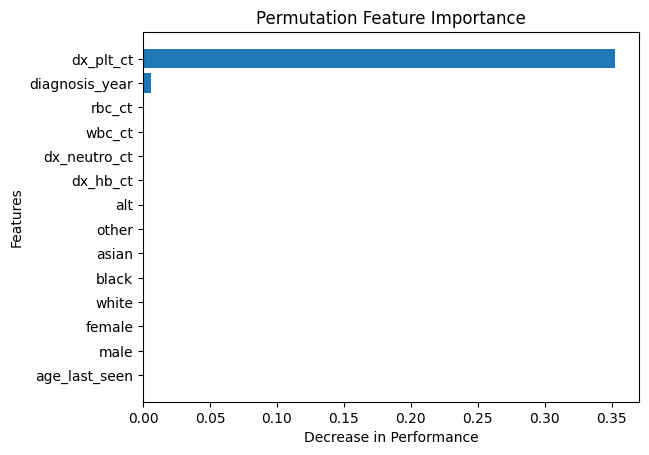}
    \includegraphics[width=.49\linewidth]{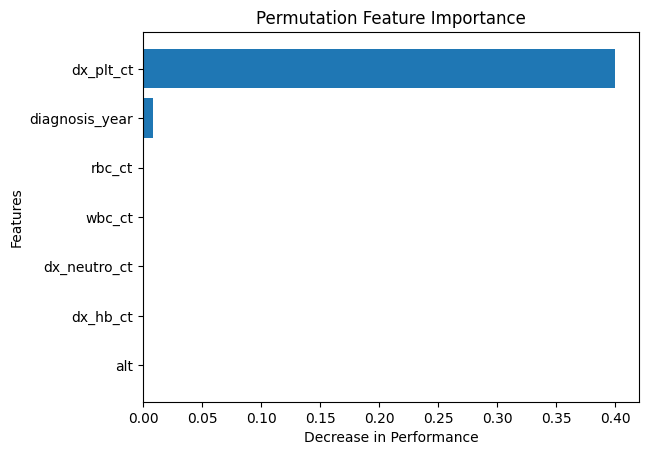}
  \caption{Permutation feature importance on the training set for the RF  model in the case of the demographic-aware (on the left side) and  demographic-unaware (on the right side) approach}
  \label{rf}
\end{figure}

\begin{figure}[h!]
\centering
  \includegraphics[width=.49\linewidth]{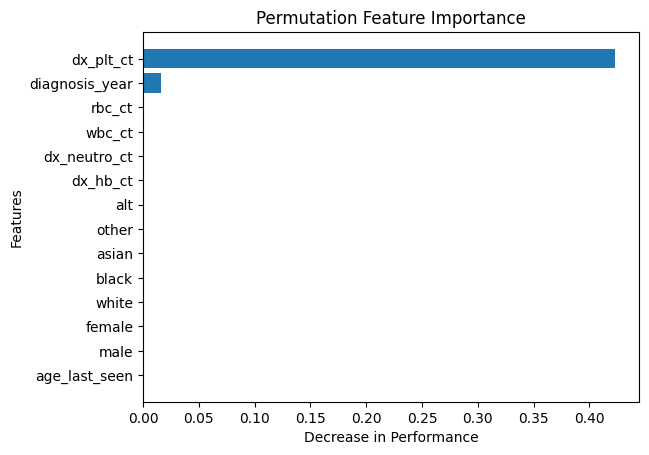}
    \includegraphics[width=.49\linewidth]{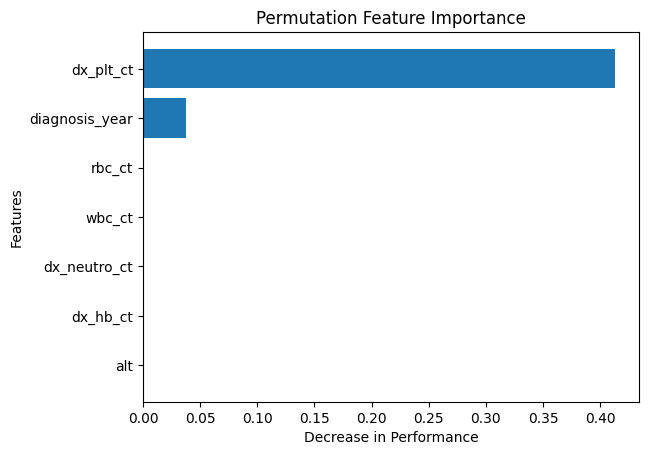}
  \caption{Permutation feature importance on the test set for the RF  model in the case of the demographic-aware (on the left side) and  demographic-unaware (on the right side) approach}
  \label{rf_test}
\end{figure}

\begin{figure}[h!]
\centering
  \includegraphics[width=.49\linewidth]{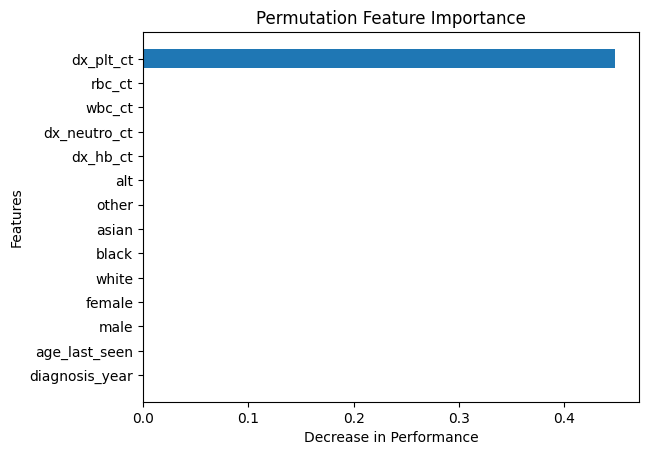}
    \includegraphics[width=.49\linewidth]{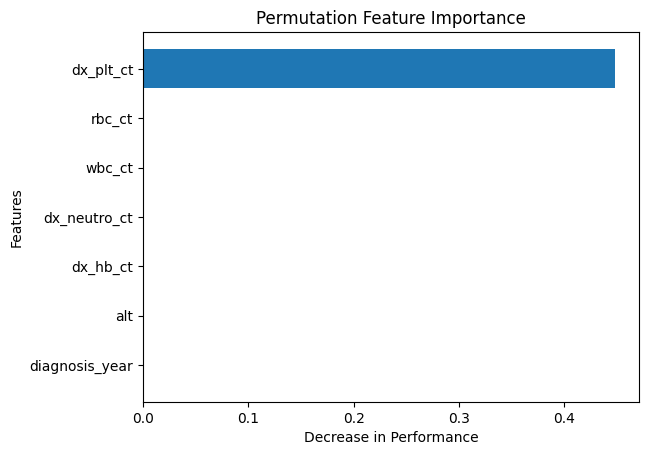}
  \caption{Permutation feature importance on the training set for the DT model in the case of the demographic-aware (on the left side) and  demographic-unaware (on the right side) approach}
  \label{dt}
\end{figure}

\begin{figure}[h!]
\centering
  \includegraphics[width=.49\linewidth]{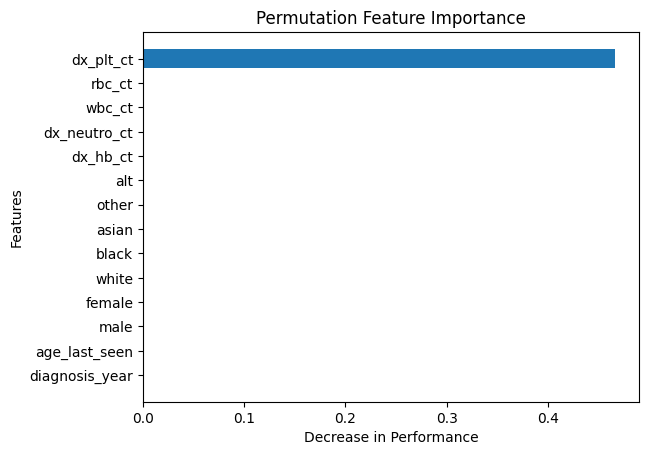}
    \includegraphics[width=.49\linewidth]{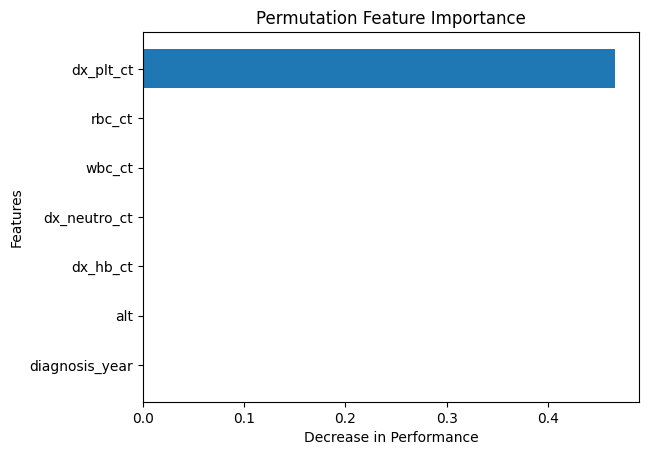}
  \caption{Permutation feature importance on the test set for the DT  model in the case of the demographic-aware (on the left side) and  demographic-unaware (on the right side) approach}
  \label{dt_test}
\end{figure}

Figures \ref{rf} and \ref{rf_test} show the permutation feature importance on the training and test sets, respectively, for the RF  model in the case of the demographic-aware (on the left side) and  demographic-unaware (on the right side) approach 
\footnote{Let us note that there were similar distributions and findings across all training and test sets of each fold (of the 5-fold cross validation); this was the case for all ML methods that we present in this subsection}
. Figures \ref{dt} and \ref{dt_test} illustrate the same information for the DT model. In all these Figures and cases, one can see that the most important and influential input variable for the ML model's decision is 'dx\_plt\_ct', i.e. the blood platelet count, regardless if we are examining the training or the test set. This is consistent with medical findings; medical experts mainly diagnose ITP based on the platelet count. Let us note that these two methods (RF and DT) are the ones that: i) achieved the best (and perfect or almost perfect) performance in diagnosing ITP (according to Table \ref{perf_comp}) compared to the other ML models; ii) are fair in terms of the sensitive variables (age, gender and race), as well as are the fairest among all ML models (according to Table \ref{fair_comp}); iii) achieved the same performance and fairness in the demographic-aware and demographic-unaware approach.

\begin{figure}[h!]
\centering
  \includegraphics[width=.49\linewidth]{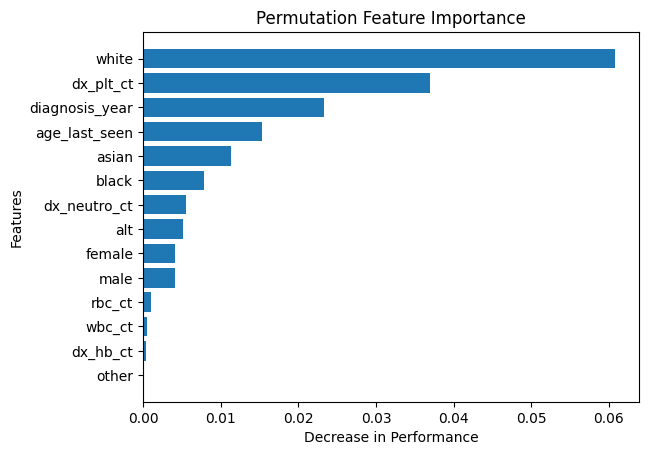}
    \includegraphics[width=.49\linewidth]{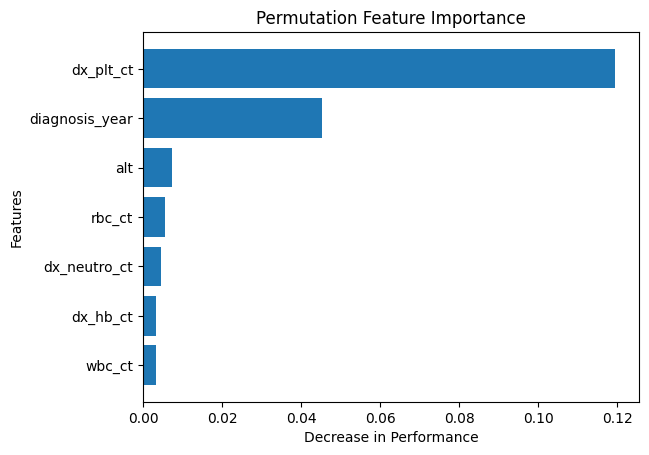}
 \caption{Permutation feature importance on the training set for the LogR model in the case of the demographic-aware (on the left side) and  demographic-unaware (on the right side) approach}
  \label{lr}
\end{figure}

\begin{figure}[h!]
\centering
  \includegraphics[width=.49\linewidth]{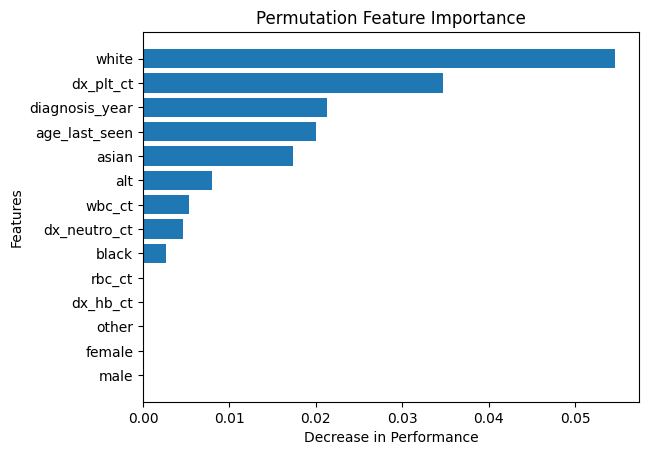}
    \includegraphics[width=.49\linewidth]{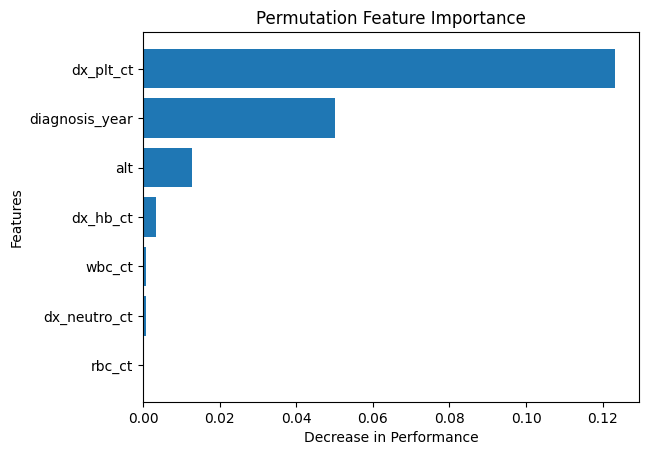}
  \caption{Permutation feature importance on the test set for the LogR model in the case of the demographic-aware (on the left side) and  demographic-unaware (on the right side) approach}
  \label{lr_test}
\end{figure}

\begin{figure}[h!]
\centering
  \includegraphics[width=.49\linewidth]{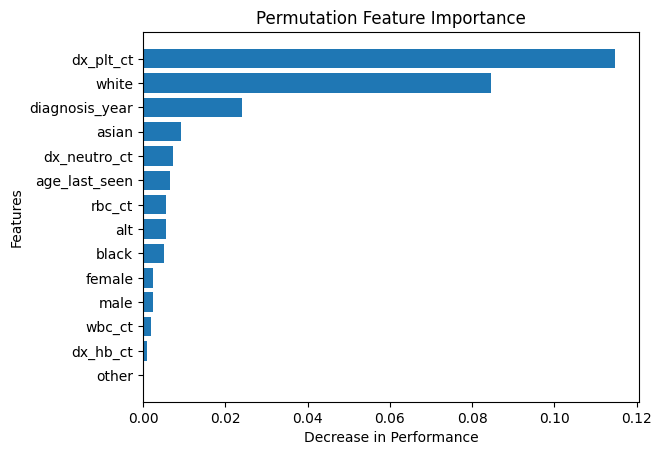}
    \includegraphics[width=.49\linewidth]{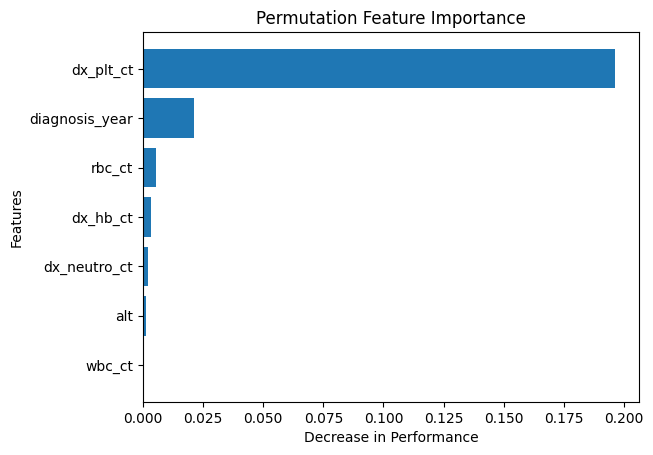}
  \caption{Permutation feature importance on the training set for the SVM-LN model in the case of the demographic-aware (on the left side) and  demographic-unaware (on the right side) approach}
  \label{svm_ln}
\end{figure}

\begin{figure}[h!]
\centering
  \includegraphics[width=.49\linewidth]{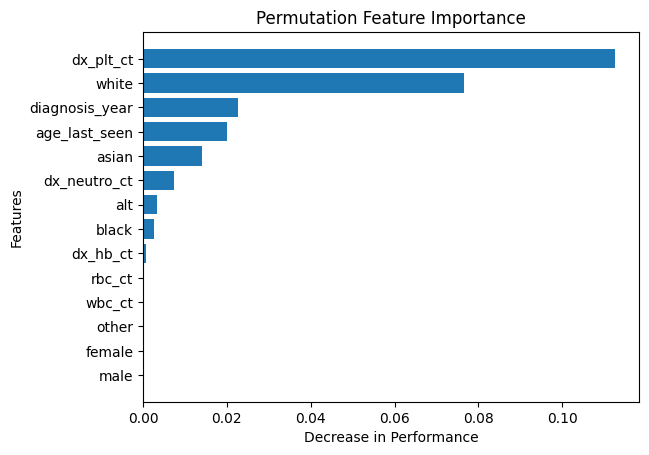}
    \includegraphics[width=.49\linewidth]{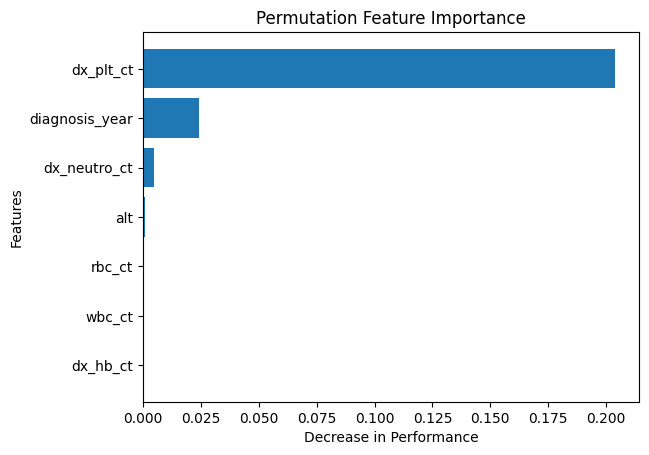}
 \caption{Permutation feature importance on the test set for the SVM-LN model in the case of the demographic-aware (on the left side) and  demographic-unaware (on the right side) approach}  \label{svm_ln_test}
\end{figure}

\begin{figure}[h!]
\centering
  \includegraphics[width=.49\linewidth]{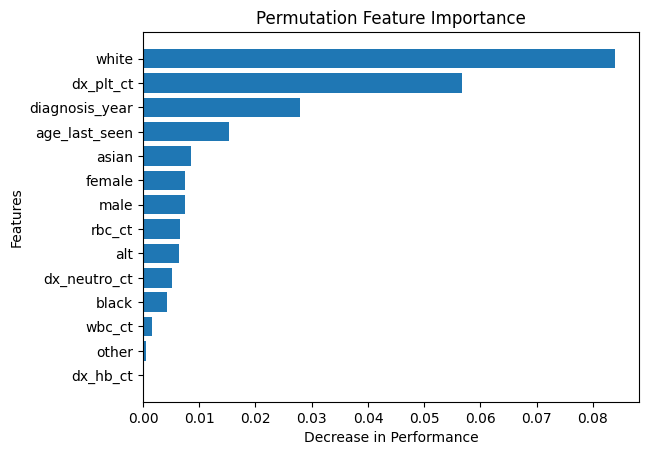}
    \includegraphics[width=.49\linewidth]{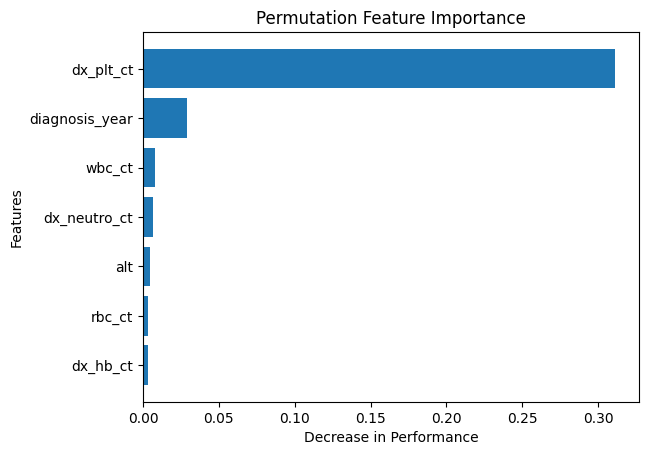}
  \caption{Permutation feature importance on the training set for the SVM-RBF model in the case of the demographic-aware (on the left side) and  demographic-unaware (on the right side) approach}
  \label{svm_rbf}
\end{figure}

\begin{figure}[h!]
\centering
  \includegraphics[width=.49\linewidth]{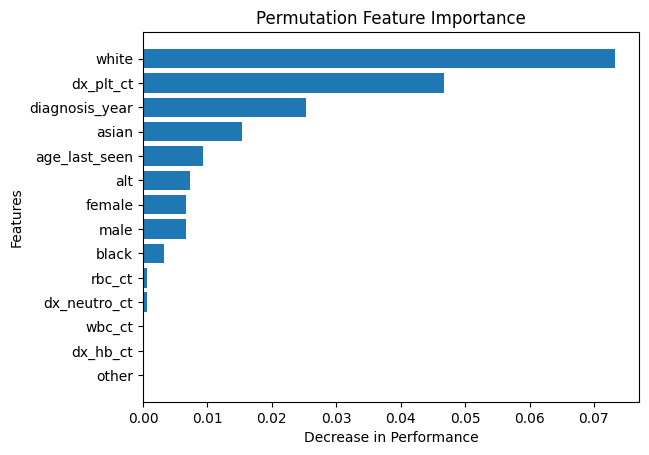}
    \includegraphics[width=.49\linewidth]{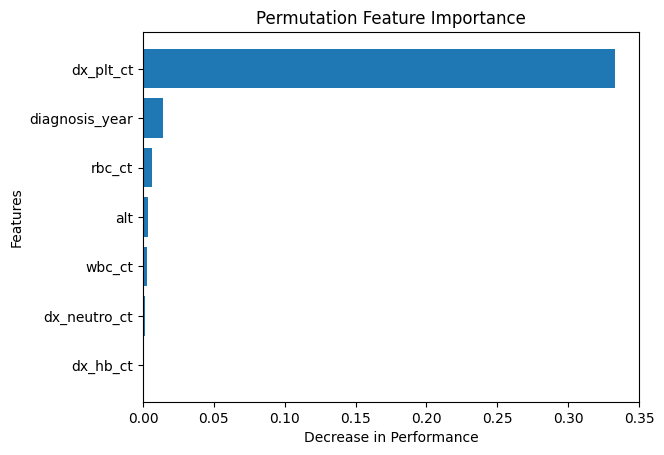}
  \caption{Permutation feature importance on the test set for the SVM-RBF model in the case of the demographic-aware (on the left side) and  demographic-unaware (on the right side) approach}
  \label{svm_rbf_test}
\end{figure}

\begin{figure}[h!]
\centering
  \includegraphics[width=.49\linewidth]{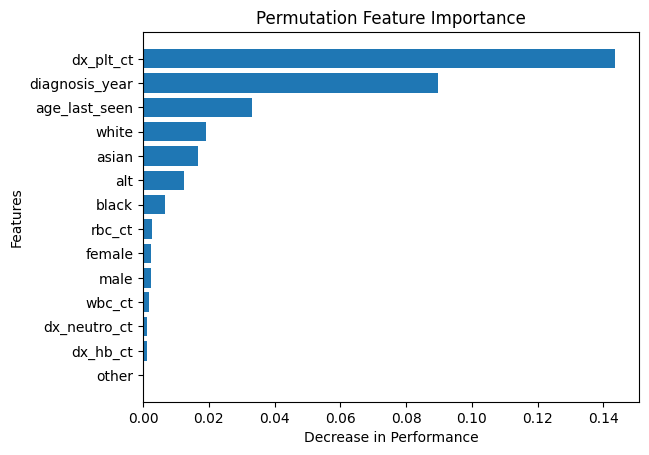}
    \includegraphics[width=.49\linewidth]{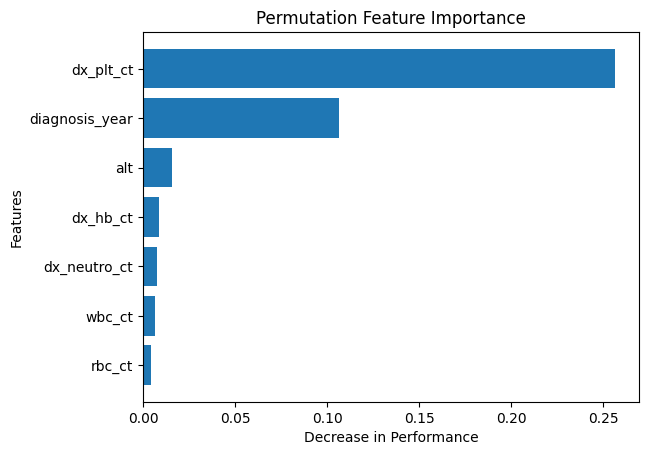}
  \caption{Permutation feature importance on the training set for the 2-NN model in the case of the demographic-aware (on the left side) and  demographic-unaware (on the right side) approach}
  \label{2nn}
\end{figure}

\begin{figure}[h!]
\centering
  \includegraphics[width=.49\linewidth]{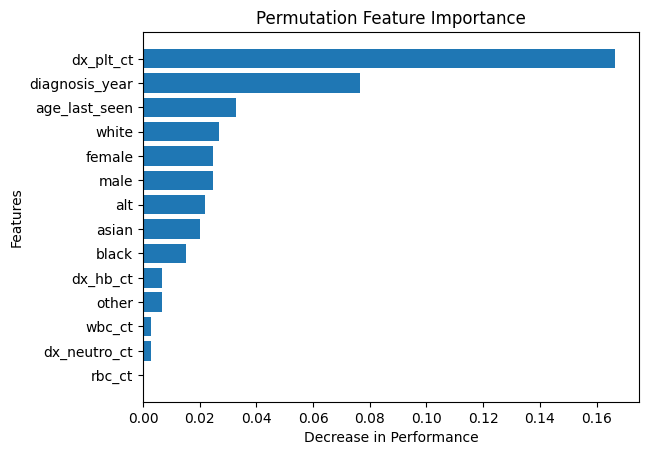}
    \includegraphics[width=.49\linewidth]{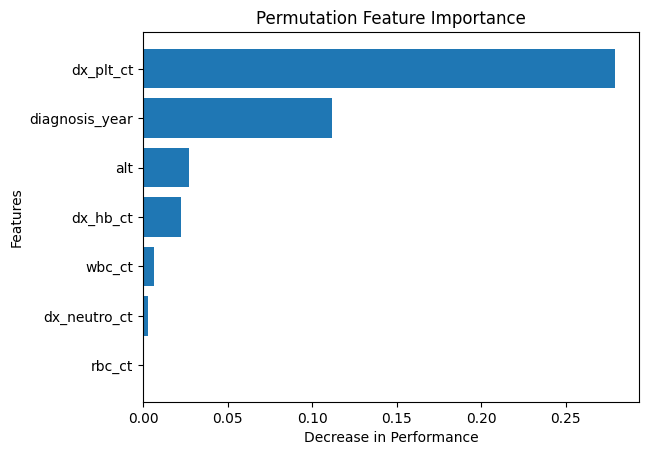}
  \caption{Permutation feature importance on the test set for the 2-NN model in the case of the demographic-aware (on the left side) and  demographic-unaware (on the right side) approach}
  \label{2nn_test}
\end{figure}

\begin{figure}[h!]
\centering
  \includegraphics[width=.49\linewidth]{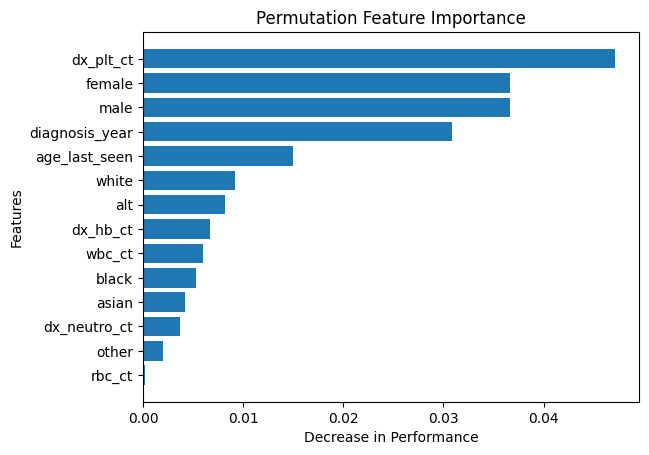}
    \includegraphics[width=.49\linewidth]{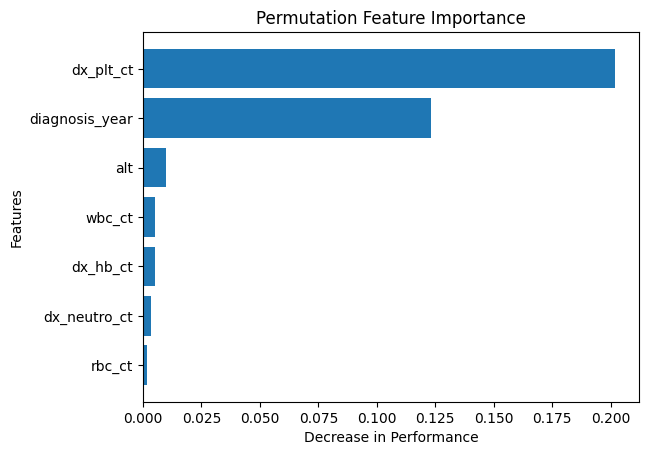}
  \caption{Permutation feature importance on the training set for the 12-NN model in the case of the demographic-aware (on the left side) and  demographic-unaware (on the right side) approach}
  \label{12nn}
\end{figure}

\begin{figure}[h!]
\centering
  \includegraphics[width=.49\linewidth]{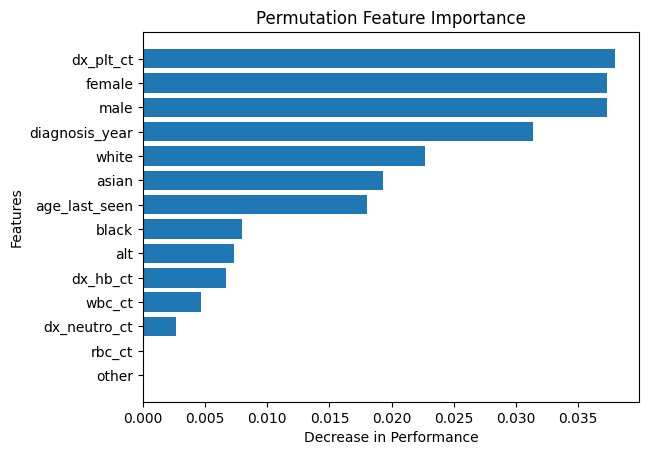}
    \includegraphics[width=.49\linewidth]{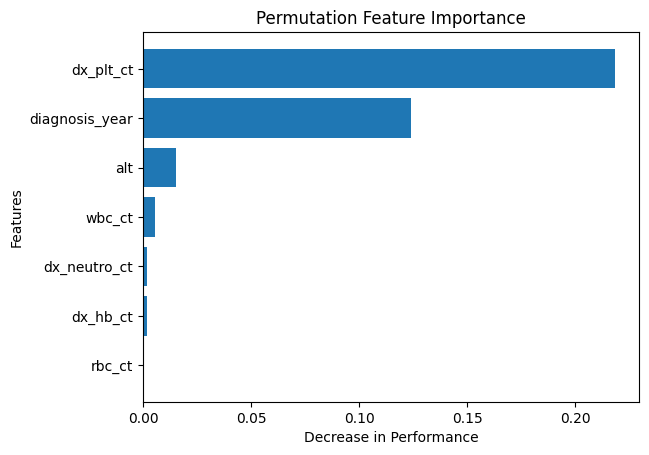}
  \caption{Permutation feature importance on the test set for the 12-NN model in the case of the demographic-aware (on the left side) and  demographic-unaware (on the right side) approach}
  \label{12nn_test}
\end{figure}

Similarly, Figures \ref{lr} (and \ref{lr_test}), \ref{svm_ln} (and \ref{svm_ln_test}), \ref{svm_rbf} (and \ref{svm_rbf_test}),  \ref{2nn} (and \ref{2nn_test}), \ref{12nn} (and \ref{12nn_test}) show the permutation feature importance on the training set (and test set) for the LogR, SMV-LN, SMV-RBF, 2-NN, 12-NN models, respectively, in the case of the demographic-aware (on the left side) and  demographic-unaware (on the right side) approach. In all cases, one can see that 'dx\_plt\_ct' is in the top-2 most important variables.

These Figures prove that, for the demographic-aware approach, the sensitive variables play an important role in the models' decisions. In all ML models, mainly the race and age are always in the top-2 most important input variables that cause the most significant decrease in performance. They also show that, for all ML models, the input variable 'dx\_plt\_ct' (i.e. the blood platelet count) plays a more important role in the demographic-unaware approach compared to the demographic-aware one; in the case of the demographic-unaware approach, the decrease in model's performance is between two and seven times bigger than the corresponding one of the demographic-aware approach (for instance, for SVM-RBF, for 'dx\_plt\_ct', the decrease in performance is around 0.06 in the  demographic-aware approach and 0.30 in the demographic-unaware approach).

Let us also note that from Table \ref{perf_comp} 12-NN (in the demographic-aware approach) and LogR (in both approaches) were the worst performing models. This is explained in Figures \ref{lr} (and \ref{lr_test}) and \ref{12nn} (and \ref{12nn_test})  as 'dx\_plt\_ct' exhibits the smallest decrease in performance (around 0.04 for the demographic-aware and 0.12 for the demographic-unaware approach) among all ML models.

%\begin{figure}[h!]
%\centering
%  \includegraphics[width=.49\linewidth]{svm_with_poly2.png}
%    \includegraphics[width=.49\linewidth]{svm_no_poly2.png}
%  \caption{SVM-P2 with (left) and without (right) demo; similar distributions across all folds; train}
%  \label{svm}
%\end{figure}

%\begin{figure}[h!]
%\centering
%  \includegraphics[width=.49\linewidth]{svm_with_test_poly2.png}
%    \includegraphics[width=.49\linewidth]{svm_no_test_poly2.png}
%  \caption{SVM-P2 with (left) and without (right) demo; similar distributions across all folds; test}
%  \label{svm}
%\end{figure}

%%%%%%%%%%%%%%%%%%%%%%%%%%%%%%%%%%%%%%%%%%
\section{Conclusions}

%\cite{cooper2019immune}

%Clinical and laboratory features present at the time of initial ITP diagnosis can be utilized to predict the development of cITP in pediatric patients using ML models. Ensemble decision tree methods are promising candidates for further ML method refinement, as AUC ROC of predicting cITP with a 100 tree RF model is > 0.7. Our group is expanding this model through incorporation of genotyping data from both acute and cITP patients. Ultimately, these ML models, in the form of an online tool, could be applied to predict cITP, allowing providers to initiate upfront interventions for those ITP patients who are unlikely to experience spontaneous disease resolution.

In conclusion, this feasibility study demonstrates the potential of ML models to significantly enhance the diagnostic process for Primary Immune Thrombocytopenia (ITP) in non-acute outpatient settings. By analyzing routine blood tests and demographic information, models such as the Random Forest and Decision Tree were found to provide high predictive performance and fairness, performing robustly across different subsets of data. The results indicate that, while models that are not presented with demographic information often achieved higher predictive performance, those presented with demographic information showed higher fairness, highlighting the complex balance between model performance and fairness. Importantly, this study identified platelet count as the most critical predictor of ITP, confirming the relevance of this parameter in clinical diagnostics. By facilitating earlier and more accurate diagnosis, the implementation of such ML models could lead to better patient management and potentially reduce the healthcare system burden associated with ITP. 

Future work includes expanding this study by incorporating a bigger data corpus (bigger in terms of ITP patients and non-ITP patients), exploring more input variables, including more conditions, causes of ITP and diseases in the non-ITP cases.

%- demographic-unaware approach tends to perform better or equally (in terms of f1 score), but is less fairer or of equal fairness (in terms of equalized odds), across all machine learning models, compared to the demographic-aware approach.: gender

%- RF and DT achieved the best performance in terms of f1 score; they achieved the same results for both the demographic-aware and the demographic-unaware approach; they achieved the best fairness performance for gender and they were fair models as they achieved 90\% or more; the rest of the models worce performance k more biased/less fair

%\textcolor{blue}{A conclusion from this and the previous subsection is that the RF and DT methods were fair models, achieved the highest fairness scores among all ML models and also achieved the same fairness score across the demographic-aware and demographic-unaware approaches. In parallel, these models achieved the highest performance across all ML models and also achieved the same performance across the two approaches. These observations mean that these models’ fairness does not change with demographic information, nor does their performance. This means that demographic features do not distract the model from focusing on more predictive features; it also means that the models do not overfit the demographic features at the expense of others that are more generalizable; finally, it means that the models remain fair w.r.t to the demographic features. In reality, these sensitive variables do not play an important role in diagnosing ITP and this is verified by medical experts.}

\vspace{6pt}

%%%%%%%%%%%%%%%%%%%%%%%%%%%%%%%%%%%%%%%%%%
\authorcontributions{
Conceptualization and methodology, D. Kollias, H. Miah, G.L. Pedone, D. Provan, F. Chen; 
software, validation, investigation and formal analysis, H. Miah, D. Kollias, G.L. Pedone, D. Provan, F. Chen; 
writing (original draft preparation, review, editing),D. Kollias, H. Miah, G.L. Pedone, D. Provan, F. Chen; 
supervision and project administration, H. Miah, D. Kollias, G.L. Pedone, D. Provan, F. Chen. All authors have read and agreed to the published version of the manuscript.
%For research articles with several authors, a short paragraph specifying their individual contributions must be provided. The following statements should be used ``Conceptualization, X.X. and Y.Y.; methodology, X.X.; software, X.X.; validation, X.X., Y.Y. and Z.Z.; formal analysis, X.X.; investigation, X.X.; resources, X.X.; data curation, X.X.; writing---original draft preparation, X.X.; writing---review and editing, X.X.; visualization, X.X.; supervision, X.X.; project administration, X.X.; funding acquisition, Y.Y. All authors have read and agreed to the published version of the manuscript.'', please turn to the  \href{http://img.mdpi.org/data/contributor-role-instruction.pdf}{CRediT taxonomy} for the term explanation. Authorship must be limited to those who have contributed substantially to the work~reported.
}

\funding{This research received no external funding.
%Please add: ``This research received no external funding'' or ``This research was funded by NAME OF FUNDER grant number XXX.'' and  and ``The APC was funded by XXX''. Check carefully that the details given are accurate and use the standard spelling of funding agency names at \url{https://search.crossref.org/funding}, any errors may affect your future funding.
}

\institutionalreview{The study was conducted in accordance with the Declaration of Helsinki, and approved by the Research Ethics Committee of QMUL and Barts Health NHS Trust (reference number 07/H0718/57, date of approval: 10/10/2023).
%In this section, you should add the Institutional Review Board Statement and approval number, if relevant to your study. You might choose to exclude this statement if the study did not require ethical approval. Please note that the Editorial Office might ask you for further information. Please add “The study was conducted in accordance with the Declaration of Helsinki, and approved by the Institutional Review Board (or Ethics Committee) of NAME OF INSTITUTE (protocol code XXX and date of approval).” for studies involving humans. OR “The animal study protocol was approved by the Institutional Review Board (or Ethics Committee) of NAME OF INSTITUTE (protocol code XXX and date of approval).” for studies involving animals. OR “Ethical review and approval were waived for this study due to REASON (please provide a detailed justification).” OR “Not applicable” for studies not involving humans or animals.
}

\informedconsent{The dataset used in this study originates from the United Kingdom Adult ITP Registry, hosted jointly by QMUL and Barts Health NHS Trust (Research Ethics Committee reference number 07/H0718/57). Subjects are de-identifiable. 
%Any research article describing a study involving humans should contain this statement. Please add ``Informed consent was obtained from all subjects involved in the study.'' OR ``Patient consent was waived due to REASON (please provide a detailed justification).'' OR ``Not applicable'' for studies not involving humans. You might also choose to exclude this statement if the study did not involve humans. Written informed consent for publication must be obtained from participating patients who can be identified (including by the patients themselves). Please state ``Written informed consent has been obtained from the patient(s) to publish this paper'' if applicable.
}

\dataavailability{Data can be available upon request from the corresponding authors.
%We encourage all authors of articles published in MDPI journals to share their research data. In this section, please provide details regarding where data supporting reported results can be found, including links to publicly archived datasets analyzed or generated during the study. Where no new data were created, or where data is unavailable due to privacy or ethical restrictions, a statement is still required. Suggested Data Availability Statements are available in section ``MDPI Research Data Policies'' at \url{https://www.mdpi.com/ethics}.
}

\conflictsofinterest{The authors declare no conflicts of interest.
}

%%%%%%%%%%%%%%%%%%%%%%%%%%%%%%%%%%%%%%%%%%
\begin{adjustwidth}{-\extralength}{0cm}

\reftitle{References}

\bibliography{references}

\PublishersNote{}
\end{adjustwidth}
\end{document}